\pdfoutput=1

\documentclass[11pt]{article}

\usepackage[]{acl}

\usepackage{times}
\usepackage{latexsym}
\usepackage{soul}
\usepackage{url}
\usepackage[utf8]{inputenc}
\usepackage{graphicx}
\usepackage{amsmath}
\usepackage{amsthm}
\usepackage{booktabs}
\usepackage{algorithm}
\usepackage{algorithmic}
\usepackage{array}
\usepackage{color, colortbl}
\usepackage{multirow}
\usepackage{setspace}
\usepackage{autobreak}
\usepackage{amsfonts}
\usepackage{amssymb}
\usepackage{amsthm,amsmath}
\usepackage{mathrsfs}
\usepackage{multirow}
\usepackage{subfigure}
\usepackage{makecell}
\usepackage{pifont}
\usepackage{fontawesome}
\usepackage{bm}

\usepackage{hyperref}

\usepackage[T1]{fontenc}

\usepackage[utf8]{inputenc}

\usepackage{microtype}

\title{Decorrelate Irrelevant, Purify Relevant: \\ Overcome Textual Spurious Correlations from a Feature Perspective}

\author{
    { Shihan Dou$^{1*}$, \ \ Rui Zheng$^{1}$\thanks{{ }{ } Equal contribution.} , \ \ Ting Wu$^{1}$, \ \ Songyang Gao$^{1}$, \ \ Junjie Shan$^{3}$,}\\
    { \bf Qi Zhang$^{12}$, \ \ Yueming Wu$^{4}$, \ \ Xuanjing Huang$^{1}$}\thanks{{ }{ }{ }Corresponding author.} \\
    {$^1$  School of Computer Science, Fudan University, Shanghai, China} \\
    {$^2$  Shanghai Key Laboratory of Intelligent Information Processing, Fudan University} \\
    {$^3$  KTH Royal Institute of Technology, Stockholm, Sweden} \\
    {$^4$  Nanyang Technological University, Singapore} \\
    \texttt{ \{shdou21,tingwu21,gaosy21\}@m.fudan.edu.cn}\\
    \texttt{ \{rzheng20,qz,xjhuang\}@fudan.edu.cn}\\
}

\begin{document}

\maketitle

\begin{abstract}

Natural language understanding (NLU) models tend to rely on spurious correlations (\emph{i.e.}, dataset bias) to achieve high performance on in-distribution datasets but poor performance on out-of-distribution ones.
Most of the existing debiasing methods often identify and weaken these samples with biased features (\emph{i.e.}, superficial surface features that cause such spurious correlations).
However, down-weighting these samples obstructs the model in learning from the non-biased parts of these samples.
To tackle this challenge, in this paper, we propose to eliminate spurious correlations in a fine-grained manner from a feature space perspective.
Specifically, we introduce Random Fourier Features and weighted re-sampling to decorrelate the dependencies between features to mitigate spurious correlations.
After obtaining decorrelated features, we further design a mutual-information-based method to purify them, which forces the model to learn features that are more relevant to tasks.
Extensive experiments on two well-studied NLU tasks demonstrate that our method is superior to other comparative approaches.
\end{abstract}

\section{Introduction}

Recently, researchers have found that the main reason why large-scale pre-trained language models perform well on NLU tasks is that they rely on \emph{spurious correlations}, rather than capturing the language understanding for the intended task \cite{bender2020climbing}.
These spurious correlations are also denoted as \emph{dataset bias} in previous work \cite{he2019unlearn,clark2019don}: prediction rules that work for training examples but do not hold in general.
In reality, a variety of spurious correlations appear in widely-used NLU benchmark datasets.
For example, in natural language inference (NLI) tasks, \citet{mccoy2019right} observe that models on the MNLI dataset \cite{williams2018broad} rely heavily on the features of word overlap to predict the entailment label blindly.
Consequently, these models perform poorly on out-of-distribution (OOD) datasets where such correlations no longer hold \cite{nie2019analyzing}.

To mitigate these spurious correlations, some existing debiasing works \cite{clark2019don,he2019unlearn} prefer to train a \emph{bias model} with known spurious correlations as prior knowledge to identify the samples without biased features.
This trained \emph{bias model} is used in the later stage to force the \emph{main model} to learn from these samples.
For better transferability, \citet{utama2020towards,sanh2020learning} relax this basic assumption that spurious correlation is apriori by using a small part of the training dataset in the training phase of \emph{bias model}.
However, these methods are not end-to-end and their training procedures are complicated.
Moreover, not all features in the samples with biased features are insignificant \cite{wen2021debiased}.
These samples may still contain features that generalize to the real-world dataset, and weakening these samples obstructs the model in learning from the non-biased parts of these samples \cite{wen2021debiased}.

In this paper, unlike the above-mentioned methods, we propose an end-to-end method that can eliminate the spurious correlations in a fine-grained way\footnote{Our code is available at \href{ https://github.com/Coling2022-DePro/DePro}{https://github.com/Coling2022-DePro/DePro}.}. Recently, some works \cite{marcus2018deep,arjovsky2019invariant} have demonstrated that spurious correlations are essentially caused by the subtle dependencies between irrelevant features (\emph{i.e.}, the features that are irrelevant to a given label) and relevant features.
According to this observation, we intend to eliminate spurious correlations by decorrelating the dependencies between features in the feature space.
However, those irrelevant features still exist in the feature space and may confuse the analysis capability of deep models.
To achieve better performance, we further design another component to purify the decorrelated features in the feature space, which forces the model to learn useful local features (\emph{i.e.}, features that are more relevant to tasks \cite{wang2020infobert}).
Specifically, we address two main challenges:
\begin{itemize}
\item \emph{Challenge 1: How to eliminate dependencies among features in the feature space?}
\item \emph{Challenge 2: How to find the useful local features and purify the decorrelated global features with them?}
\end{itemize}

To address the first challenge, some previous works \cite{shen2020stable} try to decorrelate features under linear frameworks.
However, these linear frameworks are not capable of dealing with nonlinear dependencies between features in the feature space.
To further enhance the effectiveness of methods on decorrelating nonlinear dependencies, an ideal candidate is to use kernel methods to remap the original features to high-dimensional feature space.
In this way, both linear and nonlinear dependencies can be decorrelated.
Nevertheless, the mapping operator of the kernel function cannot be given explicitly.
Therefore, we use Random Fourier Features (RFF) \cite{rahimi2007random} to approximate the kernel method for the sake of computability.
After completing high-dimensional feature reconstruction, we introduce weighted re-sampling to remove the dependencies between reconstructed features in the reconstructed feature space.
To tackle the second challenge, we introduce a saliency-map-based method to identify the useful local features in the samples and design a mutual-information-based strategy to purify the decorrelated global features (\emph{i.e.}, sentence representation) with these useful local features.

We evaluate our framework over two NLU tasks including Natural Language Inference and Fact Verification.
Through the experimental results, we observe that feature decorrelation and feature purification are both useful for improving the generalization ability of deep neural models.
Moreover, our method can achieve state-of-the-art performance on predicting out-of-distribution datasets compared with existing approaches.
In summary, this paper makes the following contributions:
\begin{itemize}
\item We introduce a novel end-to-end framework that combines feature decorrelation with feature purification to strengthen the generalization ability of NLU models.
The feature decorrelation phase is used to eliminate spurious correlations of features while the feature purification component is used to force the model to learn features that are more relevant to tasks.
\item We conduct extensive experiments over several widely used benchmark datasets. The experimental results report that feature decorrelation and feature purification can both enhance the generalization ability of deep models. Also, the results suggest the synergistic effect between decorrelation and purification. After combining them, our proposed method outperforms the state-of-the-art methods.
\end{itemize}

\section{Related Work}

\subsection{Spurious Correlations and Debiasing Methods}

\begin{figure*}[htbp]
\centerline{\includegraphics[width=1\textwidth]{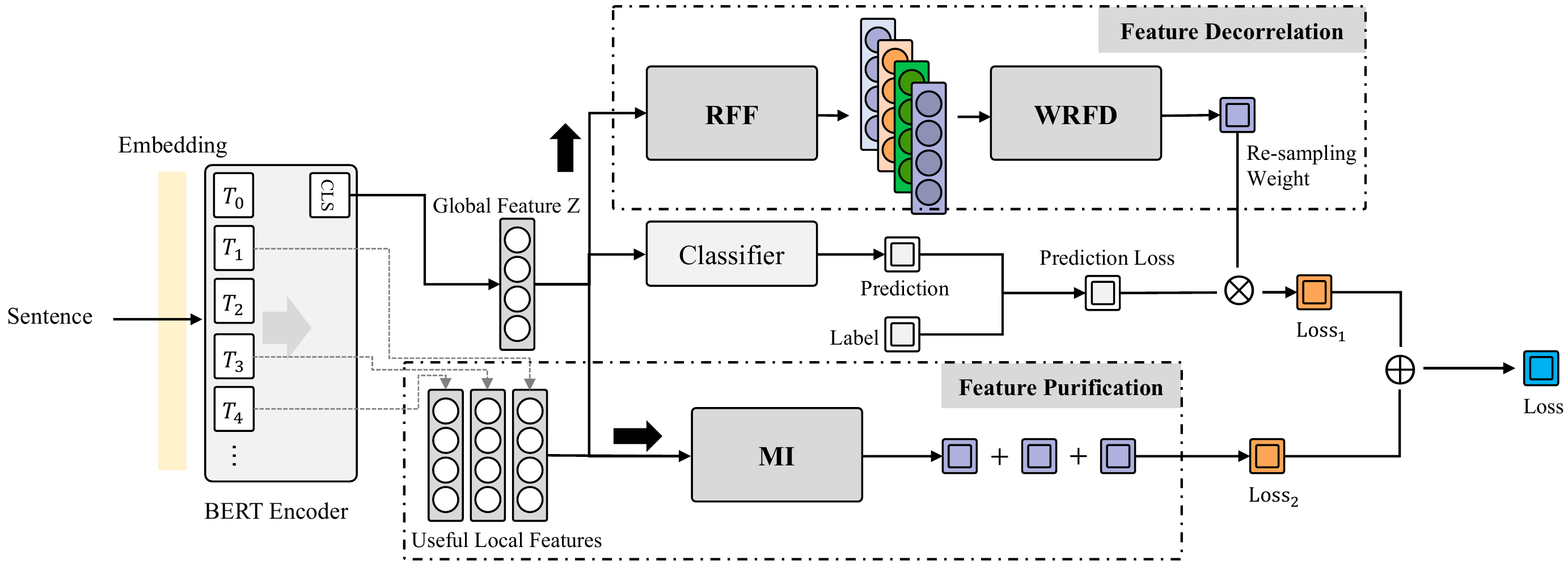}}
\caption{System architecture of \emph{DePro}. RFF, WRFD, and MI refer to Random Fourier Features, Weighted Re-sampling for Feature Decorrelation, and Mutual Information, respectively.}
\label{fig:system}
\end{figure*}

The performance of machine learning models on multiple natural language understanding benchmarks has achieved remarkable results.
However, due to the presence of spurious surface lexical-syntactic features in the training phase, deep models perform poorly on out-of-distribution examples.
These spurious properties are also known as spurious correlations or dataset biases.
For example, \citet{mccoy2019right} reports that models on the MNLI dataset \cite{williams2018broad} rely heavily on high word overlap to predict the entailment label.
In fact, spurious correlations also exist in datasets of other NLU tasks such as multi-hop QA datasets \cite{wen2021debiased}.
Deep models' excessive dependence on these spurious correlations can affect their generalization ability when testing on more challenging datasets.

In response to the problem of spurious correlations in datasets, many methods have been proposed to mitigate the impact. 
For example, \citet{clark2019don,he2019unlearn} propose a two-stage-based framework to reduce the model's dependence on known spurious correlations.
They first train a bias-only model using known spurious correlations and then leverage it to guide the main model to distinguish biased examples.
However, these approaches suffer from low transferability since they require prior knowledge about the spurious correlations in a dataset.
To mitigate the issue, \citet{utama2020towards,clark2020learning} tend to train a weak or shadow model as the bias-only model to provide guidance on discriminating biased data. 
However, these methods are not end-to-end and the training procedures of these methods are complicated.

\subsection{Feature Decorrelation}
Since the correlation between features can affect or even damage model predictions, several studies focus on eliminating this correlation during the training process.
\citet{zhang2017high} propose a strategy that selects uncorrelated features in groups to decorrelate features.
\citet{shen2020stable} address this issue by re-weighting samples. 
However, these two methods can only remove the linear dependence between features which cannot tackle the complex nonlinear dependence between features.
\citet{bahng2020learning} propose to use the biased representations to generate a debiased representation.
Although this method can decorrelate the nonlinear and linear dependence between features, it needs to artificially design the biased representation based on the known spurious correlations in the dataset.
On the contrary, our method can remove all kinds of dependencies between the features and does not need to rely on prior knowledge. 


\section{Method}

In this section, we introduce our proposed end-to-end framework namely \emph{DePro}.
Figure \ref{fig:system} presents the system architecture of \emph{DePro} which mainly consists of two phases: feature decorrelation and feature purification.
In the first phase, we introduce Random Fourier Features (RFF) \cite{rahimi2007random} to map features from the original feature space to the reconstruction space. Then we use weighted re-sampling to remove the dependencies between reconstructed features.
In the later phase, we purify the global sample features from an information theoretic perspective to further improve the generalization ability.

\textbf{Notations} 
$\mathcal{X}$, $\mathcal{Y}$, and $\mathcal{Z}$ denote the space of samples (\emph{i.e.}, sentences), the space of labels, and the feature space, respectively.
We use $f: \mathcal{X} \to \mathcal{Z}$ to denote the encoder function which can encode a sample into the feature space. The classifier function is denoted as $c : \mathcal{Z} \to \mathcal{Y}$, which can predict the sample to the corresponding label.	
Given a dataset $\mathcal{D}$ that consists of $n$ pairs of sentences and labels ${(X_i, Y_i)}_{i\in[1,n]}$, with $X_i \in \mathcal{X}$ and $Y_i \in \mathcal{Y}$, the representation of $X_i$ is denoted as $Z_i \in \mathcal{Z}$, and $Z^i$ denotes the $i$-th variable in the feature space.
For an input sentence $X_i = [X_i^1;X_i^2;\ldots;X_i^k]$, $w_i$ denotes the re-sampling weight of this sentence $X_i$ and we use $T_i = [T_i^1;T_i^2;\ldots;T_i^k]$ to denote the local feature of $X_i$ in the encoder (\emph{e.g.}, the output of BERT embedding layer).

\subsection{Decorrelate Features of Feature Space}
In this subsection, we mainly introduce our method of removing both the nonlinear and linear dependencies between features by using RFF and weighted re-sampling.

\subsubsection*{High-dimensional Feature Reconstruction via RFF}
The kernel method can obtain mutually independent features by mapping them from the original feature space to Reproducing Kernel Hilbert Space (RKHS) \cite{alvarez2012kernels} as follows:

\begin{equation}
\small
\mathcal{K}(x,\cdot)=\sum_{i=1}^{\infty} \lambda_{i} \varphi_{i}(x) \varphi_{i}(\cdot)=\left(\sqrt{\lambda}_{i} \varphi_{i}(x), \cdots\right)_{\mathcal{H}}    
\end{equation}

Here $\mathcal{K}(\cdot, \cdot)$ is the mapping operator of a measurable, symmetric positive definite kernel function and $(\cdot)_{\mathcal{H}}$ is Hilbert-Schmidt space. 
However, the mapping operator $\mathcal{K}(x,\cdot)$ is implicit. In other words, the reconstructed features cannot be obtained explicitly.
To mitigate this issue, we use Random Fourier Feature (RFF) \cite{rahimi2007random}, inspired by \citet{zhang2021deep}, to approximate the kernel function.
The function space of Random Fourier Features is denoted as $\mathcal{H}$ with the following form:
\begin{equation}
 \begin{aligned}\mathcal{H}=\{h:& x \rightarrow \sqrt{2} \cos (\omega x+\phi) \mid \\&\omega \sim N(0,1), \phi \sim \operatorname{U}(0,2 \pi)\}\end{aligned}   
\end{equation}
where $\omega$ and $\phi$ are sampled from any distribution.

For the $i$-th variable $Z^i$ and the $j$-th variable $Z^j$ of the feature space ($Z^i$ and $Z^j$ are represented by $\mathcal{A}$ and $\mathcal{B}$ for simplicity), we sample $n_{\mathcal{A}}$ and $n_{\mathcal{B}}$ mapping functions from $\mathcal{H}$ and denote them as ${u}={\{u_k\}}_{k\in [1,n_{\mathcal{A}}]}$ and ${v}={\{v_k\}}_{k \in [1,n_{\mathcal{B}}]}$.
Thus, the reconstructed features $u(\mathcal{A})$ of feature $A$ can be represented as Eq. (\ref{eq3}) and $v(\mathcal{B})$ of feature $B$ follows the same rule.
\begin{equation} \label{eq3}
\small
 \begin{array}{l}{u}(\mathcal{A})=\left(u_{1}(\mathcal{\mathcal{A}}), \ldots, u_{n_{\mathcal{A}}}(\mathcal{A})\right), u_{k}(\cdot) \in \mathcal{H}_{\text {RFF }}, \forall k, 
 \end{array}   
\end{equation}

By mapping the two features $A$ and $B$ to the reconstructed space through RFF, only linear dependencies between $u(\mathcal{A})$ and $v(\mathcal{B})$ remain.

\subsubsection*{Weighted Re-sampling for Feature Decorrelation}
We use cross-covariance operator ${\Sigma}_{XY}$ to measure the independence between features as follows:
\begin{equation}
\small
\begin{split}
 \left\langle g, \Sigma_{Y X} f\right\rangle_{\mathcal{H}_{2}}=E_{X Y}[f(X) g(Y)]- \\ E_{X}[f(X)] E_{Y}[g(Y)]
 \end{split}
\end{equation}

Specifically, for ${u}(\mathcal{A})$ and ${u}(\mathcal{B})$, the cross-covariance ${\Sigma}_{AB}$ between the distributions can be calculated by their unbiased empirical estimation with the following form:
\begin{equation} \label{eq5}
\small
  \begin{array}{r}{\Sigma}_{\mathcal{A} \mathcal{B}}=\frac{1}{n-1} \sum_{i=1}^{n}\left[\left({u}\left(\mathcal{A}_{i}\right)-\frac{1}{n} \sum_{j=1}^{n} {u}\left(\mathcal{A}_{j}\right)\right)^{T}\right.\cdot \\\left.\left({v}\left(\mathcal{B}_{i}\right)-\frac{1}{n} \sum_{j=1}^{n} {v}\left({B}_{j}\right)\right)\right]\end{array}  
\end{equation}

Hilbert-Schmidt Independence Criterion (HSIC) \cite{gretton2007kernel} uses the squared Hilbert-Schmidt norm of ${\Sigma}_{\mathcal{A}\mathcal{B}}$ to test the independence of random variables.
In the Euclidean space which the reconstructed space belongs to, Hilbert-Schmidt norm degenerates to the equivalent Frobenius norm \cite{zhang2021deep}.
Thus, we use Frobenius norm to calculate the linear correlation between the reconstructed features.


Suppose $P(\mathcal{A},\mathcal{B})$ is denoted as the joint distribution of features $\mathcal{A}$ and $\mathcal{B}$.
Due to the complicated correlation between $\mathcal{A}$ and $\mathcal{B}$, $P(\mathcal{A},\mathcal{B})$ cannot be obtained by their respective marginal distributions, which means $P(\mathcal{A}, \mathcal{B}) \neq P(\mathcal{A})\cdot P(\mathcal{B})$.
Inspired by the Acceptance-Rejection Sampling method \cite{naesseth2017reparameterization} which reparameterizes the target distribution function from the standard normal distribution by introducing the proposal distribution, we use the normalized weight function instead of the rejection process to obtain a linearly independent weighted marginal distribution from the original complex joint distribution.
Specifically, consider a probability density function with the independent marginal distributions of $\mathcal{A}$ and $\mathcal{B}$ as $\mathcal{Q}(\mathcal{A},\mathcal{B})=\mathcal{Q}(\mathcal{A})\cdot \mathcal{Q}(\mathcal{B})$, the $\mathcal{Q}$ can be fitted by the proposal distribution $\mathcal{P}$ and the normalized sampling weight is denoted as follows:
\begin{equation}
 w(x)=\frac{\mathcal{Q(x)}}{\tau \mathcal{P}(x)}
\end{equation}
where $x\in \mathcal{H} (\mathcal{A},\mathcal{B})$ and $\tau$ is a normalization constant with the following form:
\begin{equation}
 \tau^{-1} = \int_{x\in \mathcal{H}(\mathcal{A},\mathcal{ B})}^{}w(x)dx
\end{equation}
Thus, the linear dependencies between reconstructed features can be removed by the normalized weight function as follows:
\begin{equation} \label{eq8}
\small
 w(x) \cdot x(\mathcal{A},\mathcal{B})\sim \tau w(x) \cdot P(x) = \mathcal{Q}(\mathcal{A}) \cdot \mathcal{Q}(\mathcal{B})   
\end{equation}

In practice, we use the training dataset to learn the optimal sampling weights.
Through Eq. (\ref{eq5}) and Eq. (\ref{eq8}), the cross-covariance with weighted re-sampling can be estimated as:
\begin{equation}
\scriptsize
 \begin{array}{r}{\Tilde\Sigma}_{\mathcal{A} \mathcal{B};{w}}=\frac{1}{n-1} \sum_{i=1}^{n}\left[\left(w_i {u}\left(\mathcal{A}_{i}\right)-\frac{1}{n} \sum_{j=1}^{n} w_j {u}\left(\mathcal{A}_{j}\right)\right)^{T}\right.\cdot \\\left.\left(w_i {v}\left(\mathcal{B}_{i}\right)-\frac{1}{n} \sum_{j=1}^{n}  w_j {v}\left(\mathcal{B}_{j}\right)\right)\right]\end{array}
\end{equation}

As aforementioned, we use Frobenius norm to measure the correlation between features (\emph{i.e.}, $\left\|\Tilde{\Sigma}_{\mathcal{A} \mathcal{B};{w }}\right\|_{F}^{2}$).
Thus, by optimizing $w$ in the training process, both nonlinear and linear dependencies between features of the feature space can be eliminated.
Specifically, the correlation between the two variables ${Z}^i$ and ${Z}^j$ of the feature space is represented as $ \left\|\Tilde{\Sigma}_{Z^iZ^j;{w}}\right\|_{F}^{2}$.
Therefore, the re-sampling weight $w$ can be optimized as follows:
\begin{equation}
 {w}^{*}=\underset{{w} \in \mathcal{W}}{\arg \min } \sum_{1 \leq i<j \leq m_{Z}}\left\|\Tilde{\Sigma}_{{Z}^i {Z}^j ;{w}} \right\|_{F}^{2}
\end{equation}
where $\mathcal{W}=\left\{{w} \in \mathbb{R}_{+}^{n} \mid \sum_{i=1}^{n} w_{i}= n\right\}$ and $m_Z$ denotes the dimension of space $\mathcal{Z}$. 
We use a mini-batch to update the global weight repeatedly during the optimization process.
Moreover, the optimization objective function for encoder $f$ and classifier $c$ can be expressed as:
\begin{equation} \label{eq11}
  f^{*},c^{*}=\underset{f,c}{\arg \min } \sum_{i=1}^{n}w_i\mathcal{L}(c(f({X}_i)), y_i)
\end{equation}
where $\mathcal{L}(\cdot, \cdot)$ is the cross entropy loss function.				

\subsection{Feature Purification via Local Information}
For better generalization, we propose to purify the decorrelated global features from an information theoretic perspective.
Specifically, we find the useful local features by a saliency-map-based method and purify the decorrelated global features with these local features by mutual information (MI).

Inspired by \citet{han2020explaining}, we measure the significance of all local features of the sentence by computing the absolute value of the partial derivative of loss w.r.t. these local features.
The gradient of each local feature can be calculated as:
\begin{equation}
  \mathcal{G}(T^i)=\nabla_{T^i} \ell\left(f(T), y\right)  
\end{equation}
where $T^i$ is the $i$-th feature of the local features $T$.
We consider the part of the local features with the smallest values as the useless local features (\emph{e.g.}, stopwords and punctuation) which carry limited information and cannot be used to make predictions \cite{wang2020infobert}. Therefore, the information of such useless features should not be encoded into the global features of a sentence.

\begin{table*}[htbp]
  \centering
  \begin{spacing}{1.0}
    \setlength{\tabcolsep}{1.6mm}{
    
    \begin{tabular}{l|ccc|ccc}
    \toprule
    \toprule
    \multirow{2}[2]{*}{\textbf{Method}} & \multicolumn{3}{c|}{\textbf{MNLI}} & \multicolumn{3}{c}{\textbf{FEVER}} \\
          & \textbf{ID}    & \textbf{MNLI-hard} & \textbf{HANS}  & \textbf{ID}    & \textbf{Symm. v1} &
          \textbf{Symm. v2} \\
    \midrule
    BERT-base & 84.3  & 75.9  & 61.1  & 85.4  & 55.2 & 63.2 \\
    \midrule
    \rowcolor[rgb]{ .901,  .901,  .901} \multicolumn{7}{l}{\textbf{prior knowledge required}} \\
    \midrule
    Learned-Mixin + H \cite{clark2019don} & 84.2  & -     & 65.8  & 83.3  & 60.4  & 64.9 \\
    Reg-conf \cite{utama2020mind} & 84.5 & 77.3  & 69.1  & \textbf{86.4} & 60.5 & 66.2 \\
    Reweight \cite{clark2019don} & 83.5 & - & 69.2 & 84.6 & 61.7 & 66.5 \\
    PoE + CE \cite{he2019unlearn} & 83.3  & 77.6  & 67.9  & 85.7 & 57.7 & 61.4 \\
    \midrule
    \rowcolor[rgb]{ .901,  .901,  .901} \multicolumn{7}{l}{\textbf{prior knowledge NOT required}} \\
    \midrule
    MCE \cite{clark2020learning} & 83.3  & 77.6  & 64.4  & - & - & - \\
    Reg-conf \cite{utama2020towards} & 84.3  & -     & 67.1  & 87.6     & 59.8 & 66.0 \\
    PoE \cite{sanh2020learning} & 81.4  & 76.5  & 68.8  & 85.4 & 59.7 & 65.3 \\
    MoCaD \cite{xiong2021uncertainty} & 81.5  & -  & 70.0  & 87.4 & \textbf{65.7} & 69.0 \\
    \midrule
    \textbf{DePro (Our method)} & 83.2 &   \textbf{77.8}    & \textbf{70.3} & 84.5 & 65.2 & \textbf{69.2} \\
         \quad \textbf{w/o} feature decorrelation & \textbf{84.7} &   76.8    & 63.2 & 85.9 & 57.5 & 65.2 \\
         \quad \textbf{w/o} feature purification & 82.6 &   77.1   & 68.7 & 83.6 & 64.3 & 67.9 \\
    \bottomrule
    \bottomrule
    
    \end{tabular} }%
    \end{spacing}
  \caption{Accuracy results on MNLI and FEVER, and out-of-distribution test sets MNLI-hard, HANS and FEVER Symmetric (version 1\&2).
  We conduct the ablation study to further validate that our feature decorrelation and feature purification indeed improve the generalization ability.
  We compared 8 state-of-the-art debiasing methods including 4 debiasing methods with known bias and 4 debiasing methods with unknown bias.
  The hyper-parameters of BERT are identical for each model in the same dataset.
  }
  \label{tab:results}%
\end{table*}%

After feature filtering, we treat these remaining local features as useful local features that are significant to the label \cite{wang2020infobert}, and use them to purify the decorrelated sentence representation by mutual information.
Specifically, by maximizing the mutual information between the useful local features and the decorrelated sentence representation, the useful features are retained and the useless features are compressed.
In practice, we simply examine the $\ell_2$ norm of the gradient $\mathcal{G}(T^i)$ of each local feature $T^i$. 
The optimization goal can be expressed as:
\begin{equation} \label{eq13}
 \underset{f,c}{\arg \max } \ \ \alpha\sum_{j=1}^{M}I(T^j;Z)
\end{equation}
where $\alpha$ is a hyper-parameter to control the trade-off, $T^j$ is the above-mentioned useful local semantic feature, and $M$ is the number of remaining features.
In addition, due to the intractability of computing MI, we use InfoNCE \cite{oord2018representation} as the lower bound of MI to approximate $I(T^j;Z)$.

Combining Eq. (\ref{eq11}) and Eq. (\ref{eq13}), the overall optimization goal can be as follows:
\begin{equation}
\small
\begin{split}
 f^{*},c^{*}=\underset{f,c}{\arg \min } \ \ \ \sum_{i=1}^{n} (w_i \mathcal{L}(c(f({X}_i)), Y_i) - \\ \alpha\sum_{j=1}^{M}\hat{I}^{(\operatorname{InfoNCE})}(f_T(X_i^j);f({X}_i)))
 \end{split}
\end{equation}

\begin{equation}
\small
 {w}^{*}=\underset{{w} \in \mathcal{W}{n}}{\arg \min } \sum_{1 \leq i<j \leq m_{Z}}\left\|\Tilde{\Sigma}_{{Z}^i {Z}^j ;{w}} \right\|_{F}^{2}
\end{equation}
where $f_T(\cdot)$ is the function (\emph{i.e.}, the BERT embedding layer) that obtains the local features.

\section{Experiments}
In this section, we conduct extensive experiments to demonstrate (1) 
\emph{DePro} outperforms the state-of-the-art comparative approaches;
and (2) Both feature decorrelation and feature purification can improve the model's generalization ability.



\subsection{Datasets}
We conduct experiments on two well-studied NLU tasks including Natural Language Inference and Fact Verification to evaluate \emph{DePro}.
\textbf{Natural Language Inference} aims to infer the relationship between the premise and hypothesis.
For this task, we use MNLI \cite{williams2018broad} as our ID data, MNLI-hard \cite{gururangan2018annotation} and Heuristic Analysis for NLI Systems (HANS) \cite{mccoy2019right} as our OOD test set.
\textbf{Fact Verification} aims to verify a claim by the evidence document.
For this task, we use FEVER \cite{thorne2018fact} for ID evaluation and FEVER Symmetric \cite{schuster2019towards} (version 1\&2) as our OOD test set.

Specifically, we report the main results and ablation studies on the test set and evaluate all the sensitivity analyses on the development set. However, for the MNLI dataset, only the train set and dev set are publicly available, but not the published test set. So we split 10 percent of training data into a dev set dedicated to picking hyper-parameters in order to avoid overfitting. And the original dev set of MNLI is used as the test set.

\subsection{Implementation}
Similar to the majority of current debiasing methods, we choose the uncased BERT-base model \cite{devlin2018bert} as our baseline.
For all sentence-pair classification tasks, we concatenate the two sentences of one sentence pair into a single sequence and use the final-layer [CLS] embedding to represent the sentence representation.
For BERT hyper-parameters,  we use a batch size of $32$, Adam optimizer with the learning rate $5e^{-5}$ for the MNLI dataset and $2e^{-5}$ for the FEVER dataset, respectively.

For feature decorrelation, we set the learning rate of weight to $1e^{-2}$ which decays with a rate of $1e^{-3}$ for the MNLI dataset, and the learning rate to $5e^{-2}$ which decays with a rate of $1e^{-3}$ for the FEVER dataset. 
For the parameter of Random Fourier Features dimension, we have verified through extensive experiments that our method can get the best performance on HANS, Symm. v1, and Symm. v2 when the RFF dimensions are four times, two times, and four times, respectively, that of the original feature space.
For feature purification, $\alpha$ is set to $1e^{-4}$ to control the trade-off between feature decorrelation and feature purification.

\begin{table}
  \centering

  \begin{spacing}{1.0}
    \setlength{\tabcolsep}{1.4mm}{
\begin{tabular}{l|c}
\toprule
\toprule
\textbf{Method w/o Prior Knowledge} & \multicolumn{1}{l}{\textbf{End-to-End}} \\
\midrule
MCE \cite{clark2019don} & \faTimes \\
Reg-conf \cite{utama2020towards} & \faTimes \\
PoE \cite{sanh2020learning} & \faTimes \\
MoCaD \cite{xiong2021uncertainty} & \faTimes \\
\midrule
\textbf{DePro (Our method)} & \faCheckSquareO \\
\bottomrule
\bottomrule
\end{tabular}%
 }%
    \end{spacing}
  \caption{The structural details of state-of-the-art methods without the need for prior knowledge. }
  \label{tab:structural}%
\end{table}%

\subsection{Experimental Results}

\subsubsection*{Detection Performance}

Table \ref{tab:results} shows the experimental results of \emph{DePro} and comparative methods on the MNLI and FEVER datasets, respectively.
Through the table, we can see that \emph{DePro} can significantly improve the performance of the two NLU tasks and obtain state-of-the-art results on OOD datasets.
Meanwhile, the loss of \emph{DePro} on ID datasets is not significant compared to other methods.
Moreover, we also observe that the experimental results of the model under different random seeds have high variance, which has been demonstrated in previous works \cite{utama2020towards}.
To mitigate this impact, we perform our experiments with five different seeds and report the average of these results.


\begin{figure}[htbp]
\centerline{\includegraphics[width=0.47\textwidth]{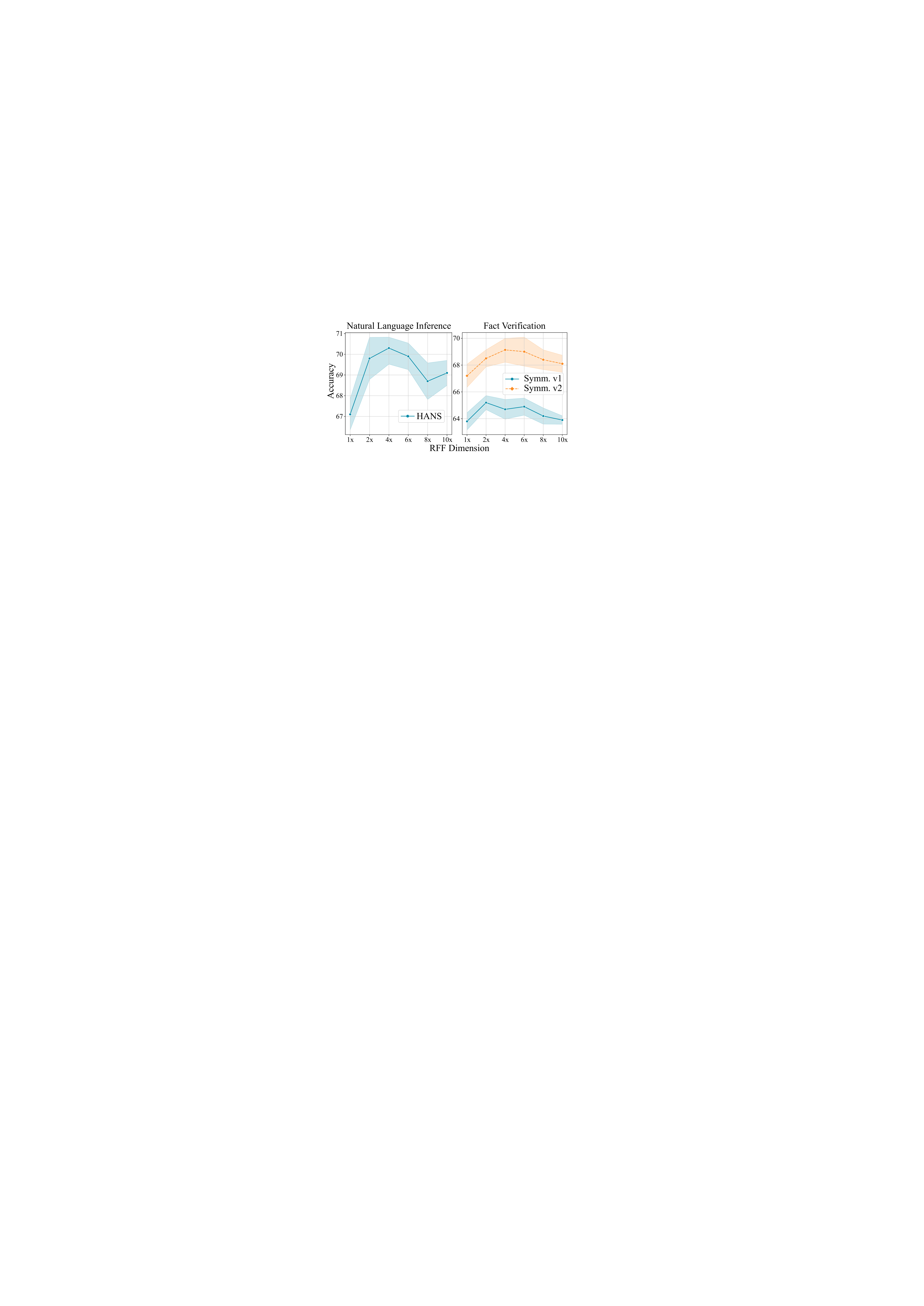}}
\caption{The results of \emph{DePro} using different RFF dimensions. Meanwhile, the ratios of feature purification for HANS, Symm. v1, and Symm. v2 are 0.7, 0.7, and 0.6, respectively.}
\label{fig:dimension}
\end{figure}

For the NLI task, compared to the baseline method (\emph{i.e.}, Uncased BERT-base model), \emph{DePro} improves by 1.9 and 9.2 percentage points on two OOD datasets MNLI-hard and HANS, respectively.
The generalization ability of \emph{DePro} on OOD datasets is also promising compared to other methods that introduce prior knowledge or unknown prior knowledge.
For the Fact Verification task, \emph{DePro} also has the best performance on the OOD dataset Symm. v2, with 6.0 percentage points higher than the accuracy of the BERT-base model.
Meanwhile, the performance of our proposed method \emph{DePro} is second only to MoCaD \cite{xiong2021uncertainty}, which is 0.5 percent lower when evaluated on Symm. v1. However, MoCaD is not an end-to-end method, but rather an improved version of the existing two-stage methods, as shown in Table \ref{tab:structural}. On the contrary, \emph{DePro} is a complete end-to-end method, which is more flexible while preserving similar detection capabilities.


In conclusion, \emph{DePro} outperforms the majority of state-of-the-art approaches on OOD datasets for two NLU tasks while the loss in ID datasets is acceptable.



\subsubsection*{Ablation Study}
\label{sec:Ablation}
We also perform two ablation experiments to check whether feature decorrelation and purification can contribute to \emph{DePro} or not.
Through the results in Table \ref{tab:results}, we find that feature decorrelation and purification can both boost the generalization ability of \emph{DePro}.
As aforementioned, the essence of spurious correlation is the subtle dependencies between relevant and irrelevant features. 
Therefore, after removing dependencies between features, we can mitigate the impact caused by spurious correlations, thus improving the model's generalization ability on OOD datasets.
The results in Table \ref{tab:results} are consistent with this situation.
On the other hand, if we directly perform feature purification on the original features, the model's performance on ID datasets can be enhanced.
It is reasonable because feature purification can align the useful local features and the sentence representation, so that the representation generated by the model is more independent of useless local features, allowing the model to focus on the useful parts of the training data.
After combining feature decorrelation with feature purification, \emph{DePro} can achieve state-of-the-art performance on distinguishing samples in OOD datasets.
Such results indicate that compared to aligning uncorrelated sentence representation, using feature purification on decorrelated representation enables sentence representation to better align the useful local features while staying away from the useless local features.

In conclusion, both feature decorrelation and feature purification can improve the detection ability, but if we can first remove the dependencies between features and then purify these decorrelated features, the generalization ability of the model can be improved to the level of state-of-the-art.


\begin{figure}[htbp]
\centerline{\includegraphics[width=0.47\textwidth]{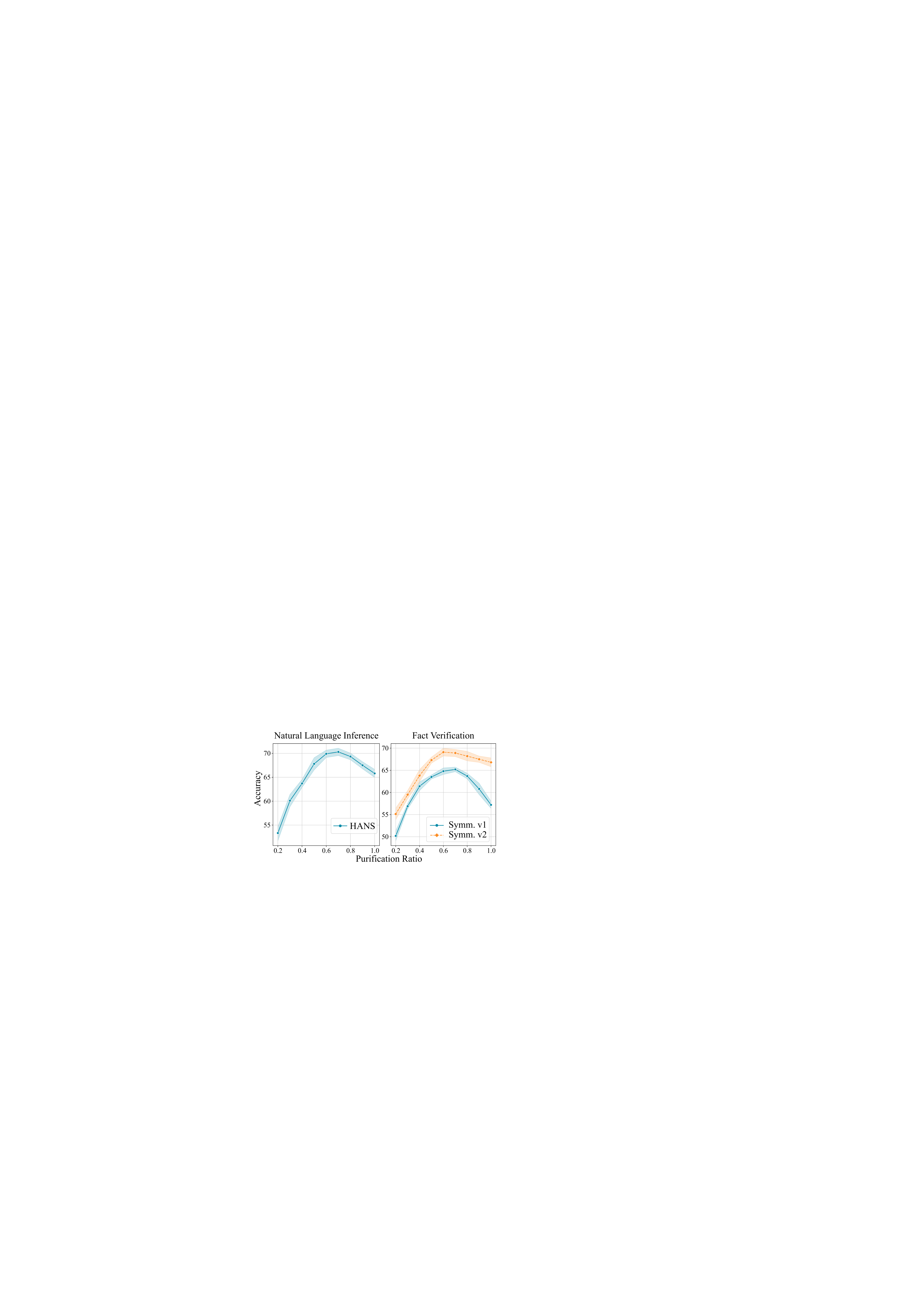}}
\caption{The results of \emph{DePro} using different purification ratios. Meanwhile, the RFF dimensions for HANS, Symm. v1, and Symm. v2 are 4x, 2x, and 4x, respectively.}
\label{fig:ratio}
\end{figure}

\subsubsection*{Sensitivity Analysis}
\label{sec:Sensitivity}

In this part, we further explore the effect of the mapping dimension size of RFF and the degree of feature purification on the generalization ability of the model.
Specifically, we choose six different RFF dimensions and nine different purification ratios to commence our study.
Due to the limited pages, we only show the corresponding experimental results of the best parameters in Figure \ref{fig:dimension} and Figure \ref{fig:ratio}.
For the NLI task, \emph{DePro} performs the best when the RFF dimension is four times that of the original features and the top 70\% of the features are used for purification.
In addition, for the FEVER dataset, \emph{DePro} can maintain the best results on Symm. v1 and Symm. v2 when the RFF dimensions are two times and four times, respectively, that of the original features and the top 70\% and 60\%, respectively, of the features are used for purification.
Through these two figures, we see that the detection effect of \emph{DePro} is different when choosing different RFF dimensions and different purification ratios.
When the dimension is expanded to a certain number, the dependencies between features can be easily removed. 
At this point, when continuing to increase the dimension, it may bring additional overhead and impact, making the detection effect decrease instead.


For feature purification, if too many local features are removed, it can make the aligned sentence representation contain too little information.
Moreover, if too many local features are purified, it may make the sentence representation contain too much useless information, so that the subsequent classifier cannot make predictions well based on the sentence representation.

\begin{figure}[htbp]
\centerline{\includegraphics[width=0.41\textwidth]{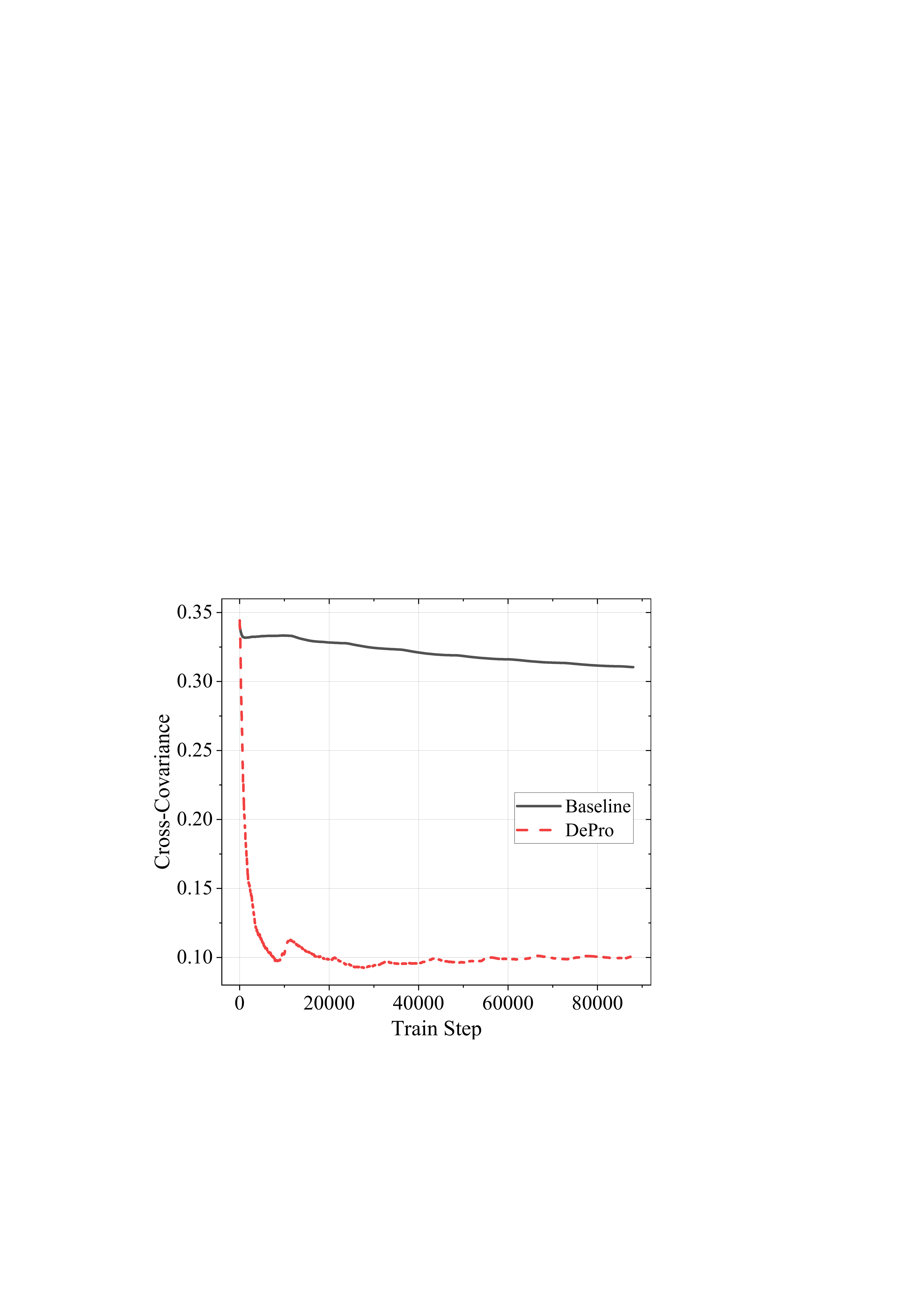}}
\caption{The mean of the correlations (\emph{i.e.}, cross-covariance) between features at different iterations.}
\label{fig:depen}
\end{figure}

\subsubsection*{Decorrelation Study}

Finally, we check whether feature decorrelation can remove the dependencies between features or not.
Specifically, during the training phase, we record the mean of the correlations between features at different iterations.
For the baseline experiment, we use the same RFF mapping functions to map the features to high-dimensional space.
However, the reconstructed features are only used to calculate the cross-covariance, not to calculate the loss and optimize the parameters.
Through the comparative results in Figure \ref{fig:depen}, we observe that the cross-covariance between features can be reduced as the number of iterations increases in \emph{DePro}.
However, in the baseline experiment, it barely decreases.


Overall, \emph{DePro} can effectively remove dependencies between features.
In this way, the spurious correlations can be mitigated at the feature level.

\begin{table}[htbp]
  \centering

  \begin{spacing}{1}
    \setlength{\tabcolsep}{1.4mm}{


\begin{tabular}{l|cc|cc}
\toprule
\toprule
\multicolumn{1}{c|}{\multirow{2}[2]{*}{\textbf{DePro }}} & \multicolumn{2}{c|}{\textbf{MNLI}} & \multicolumn{2}{c}{\textbf{FEVER}} \\
      & \textbf{ID} & \textbf{HANS} & \textbf{ID} & \textbf{Symm. v2} \\
\midrule
With $\beta$-VAE & 82.7     & 67.3     &83.6   & 65.9  \\
With RFF & \textbf{83.2}     & \textbf{70.3}     &    \textbf{84.5}   & \textbf{69.2}  \\
\bottomrule
\bottomrule
\end{tabular}%

 }%
    \end{spacing}

  \caption{Evaluation results of the feature decorrelation phase leveraging Random Fourier Features \cite{rahimi2007random} and $\beta$-VAE \cite{DBLP:conf/iclr/HigginsMPBGBML17} on two tasks, respectively.}
  \label{vaerff}%
\end{table}%


\subsection{Discussion}


In this subsection, we primarily discuss two aspects: (1) Why we choose Random Fourier Features to decorrelate features in the feature decorrelation component; and (2) What distinguishes this work from prior works that use RFF to decorrelate features. 

Many works \cite{rahimi2007random, zhang2021deep, DBLP:journals/corr/KingmaW13} have been proposed to improve the generalization of the model by performing latent representation decorrelation learning. 
We compare the performance of two decorrelation methods RFF \cite{rahimi2007random} and $\beta$-VAE \cite{DBLP:conf/iclr/HigginsMPBGBML17} in our model structure. The performance results are illustrated in Table \ref{vaerff}, which shows that RFF outperforms $\beta$-VAE in our model both in ID and OOD datasets. In contrast to RFF, VAEs decorrelate the representation while compressing it, thus damaging the generalization ability. So we choose RFF to decorrelate the feature representation to obtain the uncompressed decorrelated representation, which benefits succeeding feature purification to distinguish useful from useless local features.

The distinction between \emph{DePro} and other RFF-based methods \cite{rahimi2007random,giffon2019deep,zhang2021deep} is that our proposed method not only uses RFF for feature decorrelation but also combines two complementary approaches (\emph{i.e.}, feature decorrelation and feature purification).
These two methods are not mutually exclusive. 
In Section \ref{sec:Ablation}, we analyze the relationship between these two in detail, that is, the decorrelated features can be better purified, allowing the model to ignore more impurities when purifying useful features.
Moreover, after feature decorrelation, feature purification can constrain the model to concentrate more on useful features rather than useless features.



\section{Conclusion}
In this paper, to improve the generalization ability of deep models on OOD datasets, we design an end-to-end framework called \emph{DePro} which can eliminate spurious correlations and purify the decorrelated features. Extensive experiments on two well-studied NLU tasks demonstrate the synergistic effect between decorrelation and purification. After combining them, our method outperforms state-of-the-art methods in terms of effectiveness.

\section*{Acknowledgements}

The authors would like to thank the anonymous reviewers for their helpful comments.
We would also like to thank Feng Cheng, Haoxiang Jia, Wenxuan Li, and Yuhao Zhou for their help during the revision phase of the paper.
This work was partially National Natural Science Foundation of China (No. 62076069, 61976056), Shanghai Municipal Science and Technology Major Project (No.2021SHZDZX0103).

\bibliography{anthology,custom}
\bibliographystyle{acl_natbib}




\end{document}